\documentclass[11pt,a4paper]{article}
\usepackage[hyperref]{conll2019}

\usepackage{times}
\usepackage{latexsym}

\usepackage{url}
\usepackage{amsmath}

\usepackage{url}            %
\usepackage{booktabs}       %
\usepackage{amsfonts}       %
\usepackage{microtype}      %
\usepackage{amssymb}
\usepackage{graphicx}
\usepackage{multirow}
\usepackage{booktabs}
\usepackage{caption}
\usepackage[labelformat=empty]{subcaption} %

\usepackage{todonotes}

\usepackage{xspace}
\makeatletter
\DeclareRobustCommand\onedot{\futurelet\@let@token\@onedot}
\def\@onedot{\ifx\@let@token.\else.\null\fi\xspace}

\def\eg{\emph{e.g}\onedot}

\makeatother

\usepackage{tabularx}
\usepackage{ragged2e}
\newcolumntype{Y}{>{\RaggedRight\arraybackslash}X}
\usepackage{adjustbox}

\aclfinalcopy %

\title{Procedural Reasoning Networks for Understanding Multimodal Procedures}

\author{Mustafa Sercan Amac \quad Semih Yagcioglu \quad Aykut Erdem \quad Erkut Erdem\\
  Hacettepe University Computer Vision Lab \\
Dept. of Computer Engineering, Hacettepe University, Ankara, TURKEY  \\
 {\tt \{b21626915,n13242994,aykut,erkut\}@cs.hacettepe.edu.tr}
 }

\date{}

\begin{document}
\maketitle

\begin{abstract}
This paper addresses the problem of comprehending procedural commonsense knowledge. This is a challenging task as it requires identifying key entities, keeping track of their state changes, and understanding temporal and causal relations. Contrary to most of the previous work, in this study, we do not rely on strong inductive bias and explore the question of how multimodality can be exploited to provide a complementary semantic signal. Towards this end, we introduce a new entity-aware neural comprehension model augmented with external relational memory units. Our model learns to dynamically update entity states in relation to each other while reading the text instructions. Our experimental analysis on the visual reasoning tasks in the recently proposed RecipeQA dataset reveals that our approach improves the accuracy of the previously reported models by a large margin. Moreover, we find that our model learns effective dynamic representations of entities even though we do not use any supervision at the level of entity states.\footnote{The project website with code and demo is available at  \url{https://hucvl.github.io/prn/}}

\end{abstract}

\section{Introduction}\label{sec:intro}

\begin{figure*}[!t]
\centering
  \includegraphics[width=1\linewidth]{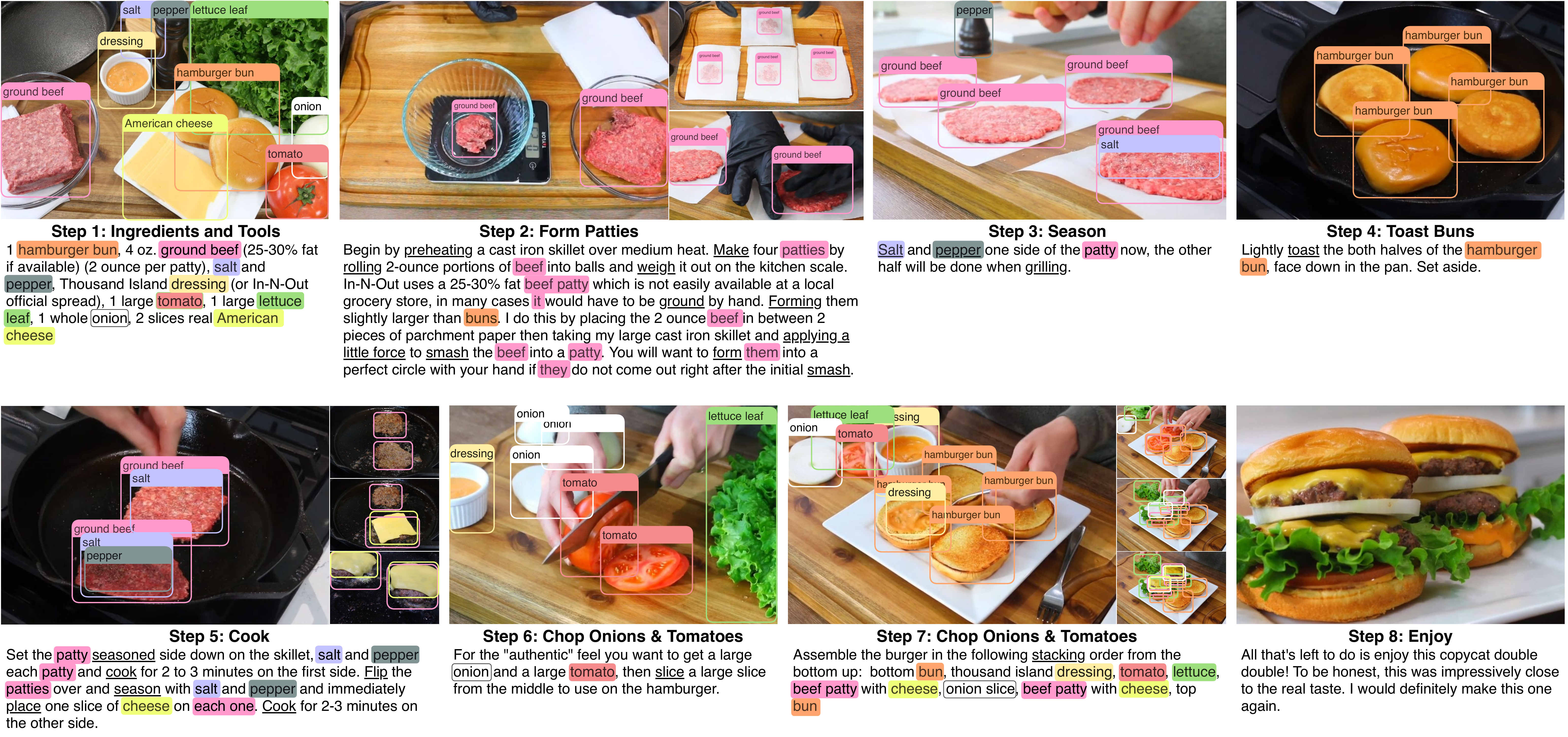}
  \caption{A recipe for preparing a cheeseburger (adapted from the cooking instructions available at {\small \url{https://www.instructables.com/id/In-N-Out-Double-Double-Cheeseburger-Copycat}}). Each basic ingredient (entity) is highlighted by a different color in the text and with bounding boxes on the accompanying images. Over the course of the recipe instructions, ingredients interact with each other, change their states by each cooking action (underlined in the text), which in turn alter the visual and physical properties of entities. For instance, the \textit{tomato} changes it form by being \underline{\textit{sliced up}} and then \underline{\textit{stacked}} on a \textit{hamburger bun}.}
  \label{fig:problem-overview}
\end{figure*}
A great deal of commonsense knowledge about the world we live is procedural in nature and involves steps that show ways to achieve specific goals. Understanding and reasoning about procedural texts (\eg cooking recipes, how-to guides, scientific processes) are very hard for machines as it demands modeling the intrinsic dynamics of the procedures ~\cite{bosselut2018simulating,mishra2018tracking,yagcioglu2018recipeqa}. That is, one must be aware of the entities present in the text, infer relations among them and even anticipate changes in the states of the entities after each action. For example, consider the cheeseburger recipe presented in Fig.~\ref{fig:problem-overview}. The instruction ``\textit{\underline{salt and pepper} each patty and \underline{cook} for 2 to 3 minutes on the first side}'' in Step 5 entails mixing three basic ingredients, the \textit{ground beef}, \textit{salt} and \textit{pepper}, together and then applying heat to the mix, which in turn causes chemical changes that alter both the appearance and the taste. From a natural language understanding perspective, the main difficulty arises when a model sees the word \textit{patty} again at a later stage of the recipe. It still corresponds to the same entity, but its form is totally different. 

Over the past few years, many new datasets and approaches have been proposed that address this inherently hard problem~\cite{bosselut2018simulating,mishra2018tracking,tandon2018reasoning,du2019consistent}. To mitigate the aforementioned challenges, the existing works rely mostly on heavy supervision and focus on predicting the individual state changes of entities at each step. Although these models can accurately learn to make local predictions, they may lack global consistency~\cite{tandon2018reasoning,du2019consistent}, not to mention that building such annotated corpora is very labor-intensive. In this work, we take a different direction and explore the problem from a multimodal standpoint. Our basic motivation, as illustrated in Fig.~\ref{fig:problem-overview}, is that accompanying images provide complementary cues about causal effects and state changes. For instance, it is quite easy to distinguish raw meat from cooked one in visual domain. 

In particular, we take advantage of recently proposed RecipeQA dataset~\cite{yagcioglu2018recipeqa}, a dataset for multimodal comprehension of cooking recipes, and ask whether it is possible to have a model which employs dynamic representations of entities in answering questions that require multimodal understanding of procedures. To this end, inspired from~\cite{santoro2018relational}, we propose Procedural Reasoning Networks (PRN) that incorporates entities into the comprehension process and allows to keep track of entities, understand their interactions and accordingly update their states across time. 
We report that our proposed approach significantly improves upon previously published results on visual reasoning tasks in RecipeQA, which test understanding causal and temporal relations from images and text. We further show that the dynamic entity representations can
capture semantics of the state information in the corresponding steps.

\section{Visual Reasoning in RecipeQA}
In our study, we particularly focus on the visual reasoning tasks of RecipeQA, namely  \emph{visual cloze}, \emph{visual coherence}, and \emph{visual ordering} tasks, each of which examines a different reasoning skill\footnote{We intentionally leave the textual cloze task out from our experiments as the questions in this task does not necessarily need multimodality.}. We briefly describe these tasks below.\\

\noindent\textbf{Visual Cloze}. In the visual cloze task, the question is formed by a sequence of four images  from consecutive steps of a recipe where one of them is replaced by a placeholder. A model should select the correct one from a multiple-choice list of four answer candidates to fill in the missing piece. In that regard, the task inherently requires aligning visual and textual information and understanding temporal relationships between the cooking actions and the entities.\\

\noindent\textbf{Visual Coherence}. The visual coherence task tests the ability to identify the  image within a sequence of four images that is inconsistent with the text instructions of a cooking recipe. To succeed in this task, a model should have a clear understanding of the procedure described in the recipe and at the same time connect language and vision.\\

\noindent\textbf{Visual Ordering}. The visual ordering task is about grasping the temporal flow of visual events with the help of the given recipe text. The questions show a set of four images from the recipe and the task is to sort jumbled images into the correct order. Here, a model needs to infer the temporal relations between the images and align them with the recipe steps.

\section{Procedural Reasoning Networks}\label{sec:model}

\begin{figure*}[!t]
\centering
  \includegraphics[width=1\linewidth]{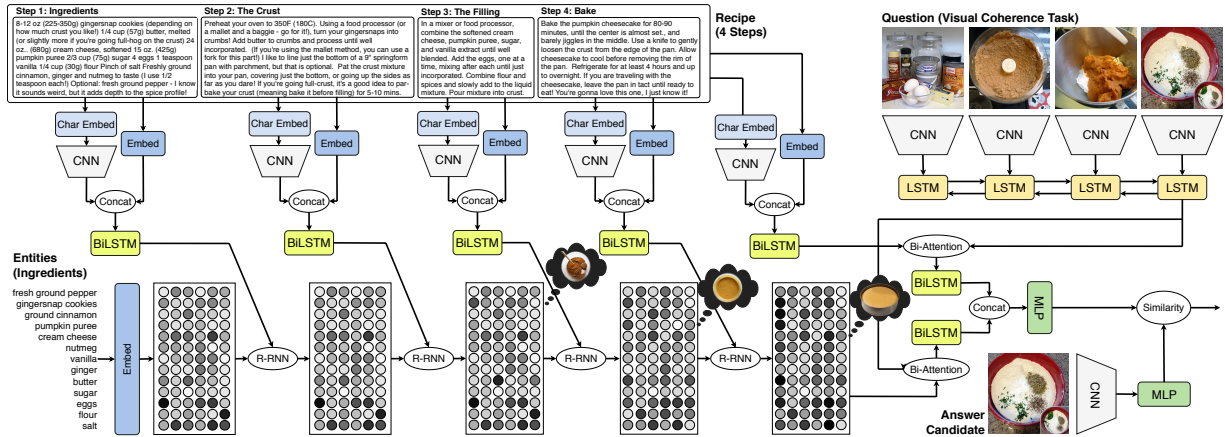}
  \caption{An illustration of our Procedural Reasoning Networks (PRN). For a sample question from visual coherence task in RecipeQA, while reading the cooking recipe, the model constantly performs updates on the representations of the entities (ingredients) after each step and makes use of their representations along with the whole recipe when it scores a candidate answer. Please refer to the main text for more details.}
  \label{fig:overview}
\end{figure*}

In the following, we explain our Procedural Reasoning Networks model. Its architecture is based on a bi-directional attention flow (BiDAF) model~\cite{gardner-2018-allennlp}\footnote{Our implementation is based on the implementation publicly available in AllenNLP~\cite{gardner-2018-allennlp}.}, but also equipped with an explicit reasoning module that acts on entity-specific relational memory units. 
Fig. \ref{fig:overview} shows an overview of the network architecture. It consists of five main modules: An input module, an attention module, a reasoning module, a modeling module, and an output module. Note that the question answering tasks we consider here are multimodal in that while the context is a procedural text, the question and the multiple choice answers are composed of images.

\begin{enumerate}
\setlength\itemsep{-0.2em}
    \item \textbf{Input Module} extracts vector representations of inputs at different levels of granularity by using several different encoders.
    \item \textbf{Reasoning Module} scans the procedural text and tracks the states of the entities and their relations through a recurrent relational memory core unit~\cite{santoro2018relational}.
    \item \textbf{Attention Module} computes context-aware query vectors and query-aware context vectors as well as query-aware memory vectors.
    \item \textbf{Modeling Module} employs two multi-layered RNNs to encode previous layers outputs.
    \item \textbf{Output Module} scores a candidate answer from the given multiple-choice list.
\end{enumerate}

At a high level, as the model is reading the cooking recipe, it continually updates the internal memory representations of the entities (ingredients) based on the content of each step -- it keeps track of changes in the states of the entities, providing an entity-centric summary of the recipe. The response to a question and a possible answer depends on the representation of the recipe text as well as the last states of the entities. All this happens in a series of implicit relational reasoning steps and there is no need for explicitly encoding the state in terms of a predefined vocabulary.

\subsection{Input Module}
Let the triple $(\mathbf{R},\mathbf{Q},\mathbf{A})$ be a sample input. Here, $\mathbf{R}$ denotes the input recipe which contains textual instructions composed of $N$ words in total. $\mathbf{Q}$ represents the question that consists of a sequence of $M$ images. $\mathbf{A}$ denotes an answer that is either a single image or a series of $L$ images depending on the reasoning task. In particular, for the visual cloze and the visual coherence type questions, the answer contains a single image ($L=1$) and for the visual ordering task, it includes a sequence.

We encode the input recipe $\mathbf{R}$ at character, word, and step levels. Character-level embedding layer uses a convolutional neural network, namely CharCNN model by \citet{kim2014convolutional}, which outputs character level embeddings for each word and alleviates the issue of out-of-vocabulary (OOV) words. In word embedding layer, we use a pretrained GloVe model \cite{pennington2014glove} and extract word-level embeddings\footnote{We also consider pretrained ELMo embeddings~\cite{peters2018deep} in our experiments but found out that the performance gain does not justify the computational overhead.}. %
The concatenation of the character and the word embeddings are then fed to a two-layer highway network~\cite{srivastava2015highway} to obtain a contextual embedding for each word in the recipe. This results in the matrix $\mathbf{R}' \in \mathbb{R}^{2d \times N}$.

On top of these layers, we have another layer that encodes the steps of the recipe in an individual manner. Specifically, we obtain a step-level contextual embedding of the input recipe containing $T$ steps as $\mathcal{S}=(\mathbf{s}_1,\mathbf{s}_2,\dots,\mathbf{s}_T)$ where $\mathbf{s}_i$ represents the final state of a BiLSTM encoding the $i$-th step of the recipe obtained from the character and word-level embeddings of the tokens exist in the corresponding step.

We represent both the question $\mathbf{Q}$ and the answer $\mathbf{A}$ in terms of visual embeddings. Here, we employ a pretrained ResNet-50 model \cite{he2016deep} trained on ImageNet dataset \cite{deng2009imagenet} and represent each image as a real-valued 2048-d vector using features from the penultimate average-pool layer. Then these embeddings are passed first to a multilayer perceptron (MLP) and then its outputs are fed to a BiLSTM. We then form a matrix $\mathbf{Q}' \in \mathbb{R}^{2d \times M}$ for the question by concatenating the cell states of the BiLSTM. For the visual ordering task, to represent the sequence of images in the answer with a single vector, we additionally use a BiLSTM and define the answering embedding by the summation of the cell states of the BiLSTM. Finally, for all tasks, these computations produce answer embeddings denoted by $\mathbf{a} \in \mathbb{R}^{2d \times 1}$.

\subsection{Reasoning Module}

\begin{figure*}[!t]
\centering
  \includegraphics[width=1\linewidth]{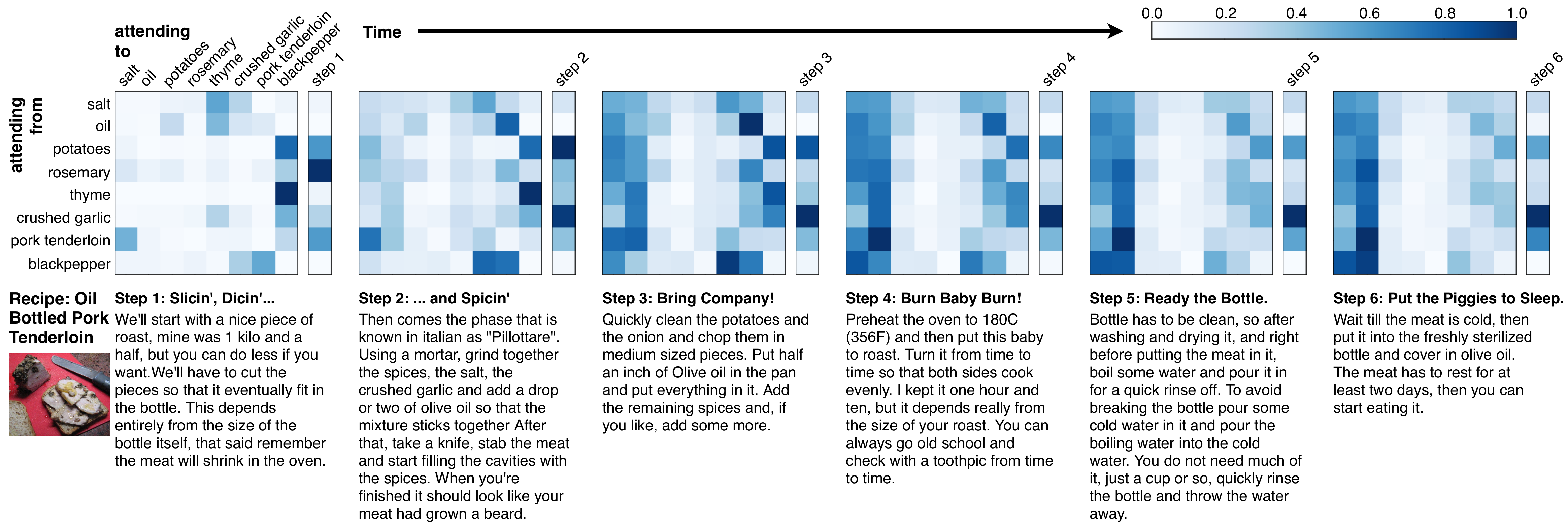}
  \caption{Sample visualizations of the self-attention weights demonstrating both the interactions among the ingredients and between the ingredients and the textual instructions throughout the steps of a sample cooking recipe from RecipeQA (darker colors imply higher attention weights). The attention maps do not change much after the third step as the steps after that mostly provide some redundant information about the completed recipe.}
  \label{fig:memoryinteraction}
\end{figure*}

As mentioned before, comprehending a cooking recipe is mostly about entities (basic ingredients) and actions (cooking activities) described in the recipe instructions. Each action leads to changes in the states of the entities, which usually affects their visual characteristics. A change rarely occurs in isolation; in most cases, the action affects multiple entities at once. Hence, in our reasoning module, we have an explicit memory component implemented with relational memory units~\cite{santoro2018relational}. This helps us to keep track of the entities, their state changes and their relations in relation to each other over the course of the recipe (see Fig. \ref{fig:memoryinteraction}). As we will examine in more detail in Section~\ref{sec:experiments}, it also greatly improves the interpretability of model outputs.

Specifically, we set up the memory with a memory matrix $\mathbf{E} \in \mathbb{R}^{d_E \times K}$ by extracting $K$ entities (ingredients) from the first step of the recipe\footnote{The first steps of the recipes in RecipeQA commonly contain a list of ingredients.}. We initialize each memory cell $\mathbf{e}_i$ representing a specific entity by its CharCNN and pre-trained GloVe embeddings\footnote{Multi-word entities (\eg \textit{minced garlic}) are represented by the average embedding vector of the words that they contain, and OOV words are expressed with the average word vector of all the words.}. From now on, we will use the terms memory cells and entities interchangeably throughout the paper. Since the input recipe is given in the form of a procedural text decomposed into a number of steps, we update the memory cells after each step, reflecting the state changes happened on the entities. This update procedure is modelled via a relational recurrent neural network (R-RNN), recently proposed by \citet{santoro2018relational}. It is built on a 2-dimensional LSTM model whose matrix of cell states represent our memory matrix $\mathbf{E}$. Here, each row $i$ of the matrix $\mathbf{E}$ refers to a specific entity $\mathbf{e}_i$ and is updated after each recipe step $t$ as follows:
\begin{equation}
    \mathbf{\phi}_{i,t}=\operatorname{R-RNN}(\mathbf{\phi}_{i,t-1}, \mathbf{s}_{t})
\end{equation}
where $\mathbf{s}_{t}$ denotes the embedding of recipe step~$t$ and  $\mathbf{\phi}_{i,t}=(\mathbf{h}_{i,t},\mathbf{e}_{i,t})$ is the cell state of the \mbox{R-RNN} at step $t$ with $\mathbf{h}_{i,t}$ and  $\mathbf{e}_{i,t}$ being the $i$-th row of the hidden state of the R-RNN and the dynamic representation of entity $\mathbf{e}_{i}$ at the step $t$, respectively. 
The R-RNN model exploits a multi-headed self-attention mechanism~\cite{vaswani2017transformer} that allows memory cells to interact with each other and attend multiple locations simultaneously during the update phase. 

In Fig. \ref{fig:memoryinteraction}, we illustrate how this interaction takes place in our relational memory module by considering a sample cooking recipe and by presenting how the attention matrix changes throughout the recipe. In particular, the attention matrix at a specific time shows the attention flow from one entity (memory cell) to another along with the attention weights to the corresponding recipe step (offset column). The color intensity shows the magnitude of the attention weights. As can be seen from the figure, the internal representations of the entities are actively updated at each step. Moreover, as argued in~\cite{santoro2018relational}, this can be interpreted as a form of relational reasoning as each update on a specific memory cell is operated in relation to others. Here, we should note that it is often difficult to make sense of these attention weights. However, we observe that the attention matrix changes very gradually near the completion of the recipe.

\subsection{Attention Module}
Attention module is in charge of linking the question with the recipe text and the entities present in the recipe. It takes the matrices $\mathbf{Q'}$ and $\mathbf{R}'$ from the input module, and $\mathbf{E}$ from the reasoning module and constructs the question-aware recipe representation $\mathbf{G}$ and the question-aware entity representation $\mathbf{Y}$. Following the attention flow mechanism described in \cite{seo2016bidirectional}, we specifically calculate attentions in four different directions: (1) from question to recipe, (2) from recipe to question, (3) from question to entities, and (4) from entities to question. 

The first two of these attentions require computing a shared affinity matrix $\mathbf{S}^R \in \mathbb{R}^{N \times M}$ with $\mathbf{S}^R_{i,j}$ indicating the similarity between $i$-th recipe word and $j$-th image in the question estimated by
\begin{equation}
    \mathbf{S}^R_{i,j}=\mathbf{w}^{\top}_{R}[\mathbf{R}'_i;\mathbf{Q}'_j;\mathbf{R}'_i \circ \mathbf{Q}'_j]
\end{equation}
where $\mathbf{w}^{\top}_{R}$ is a trainable weight vector, $\circ$ and $[;]$ denote elementwise multiplication and concatenation operations, respectively.

Recipe-to-question attention determines the images within the question that is most relevant to each word of the recipe. Let $\mathbf{\tilde{Q}} \in \mathbb{R}^{2d \times N}$ represent the recipe-to-question attention matrix with its $i$-th column being given by $ \mathbf{\tilde{Q}}_i=\sum_j \mathbf{a}_{ij}\mathbf{Q}'_j$ where the attention weight is computed by $\mathbf{a}_i=\operatorname{softmax}(\mathbf{S}^R_{i}) \in \mathbb{R}^M$.

Question-to-recipe attention signifies the words within the recipe that have the closest similarity to each image in the question, and construct an attended recipe vector given by $ \tilde{\mathbf{r}}=\sum_{i}\mathbf{b}_i\mathbf{R}'_i$ with the attention weight is calculated by $\mathbf{b}=\operatorname{softmax}(\operatorname{max}_{\mathit{col}}(\mathbf{S}^R)) \in \mathbb{R}^{N}$
where $\operatorname{max}_{\mathit{col}}$ denotes the maximum function across the column. The question-to-recipe matrix is then obtained by replicating $\tilde{\mathbf{r}}$ $N$ times across the column, giving $\tilde{\mathbf{R}} \in \mathbb{R}^{2d \times N}$.

Then, we construct the question aware representation of the input recipe, $\mathbf{G}$, with its $i$-th column $\mathbf{G}_i \in \mathbb{R}^{8d \times N}$ denoting the final embedding of $i$-th word given by
\begin{equation}
    \mathbf{G}_i =[\mathbf{R}'_i;\mathbf{\tilde{Q}}_i;\mathbf{R}'_i \circ \mathbf{\tilde{Q}}_i;\mathbf{R}'_i \circ \mathbf{\tilde{R}}_i;]\;\;.
\end{equation}

Attentions from question to entities, and from entities to question are computed in a way similar to the ones described above. The only difference is that it uses a different shared affinity matrix to be computed between the memory encoding entities~$\mathbf{E}$ and the question~$\mathbf{Q}'$. These attentions are then used to construct the question aware representation of entities, denoted by $\mathbf{Y}$, that links and integrates the images in the question and the entities in the input recipe.

\subsection{Modeling Module}
Modeling module takes the question-aware representations of the recipe $\mathbf{G}$ and the entities $\mathbf{Y}$, and forms their combined vector representation. For this purpose, we first use a two-layer BiLSTM to read the question-aware recipe $\mathbf{G}$ and to encode the interactions among the words conditioned on the question. For each direction of BiLSTM , we use its hidden state after reading the last token as its output. In the end, we obtain a vector embedding $\mathbf{c} \in \mathbb{R}^{2d \times 1}$. Similarly, we employ a second BiLSTM, this time, over the entities $\mathbf{Y}$, which results in another vector embedding $\mathbf{f} \in \mathbb{R}^{2d_E \times 1}$. Finally, these vector representations are concatenated and then projected to a fixed size representation using $\mathbf{o}=\varphi_o(\left[\mathbf{c}; \mathbf{f}\right]) \in \mathbb{R}^{2d \times 1}$ where $\varphi_o$ is a multilayer perceptron with $\operatorname{tanh}$ activation function. 

\subsection{Output Module}
The output module takes the output of the modeling module, encoding vector embeddings of the question-aware recipe and the entities $\mathbf{Y}$, and the embedding of the answer $\mathbf{A}$, and returns a similarity score which is used while determining the correct answer. Among all the candidate answer, the one having the highest similarity score is chosen as the correct answer. To train our proposed procedural reasoning network, we employ a hinge ranking loss~\citep{collobert2011jmlr}, similar to the one used in~\cite{yagcioglu2018recipeqa}, given below.
\begin{equation}
    L=\max \{0, \gamma-\cos(\mathbf{o}, \mathbf{a}_{+})+\cos(\mathbf{o}, \mathbf{a}_{-})\}
\end{equation}
where $\gamma$ is the margin parameter, $\mathbf{a}_+$ and $\mathbf{a}_{-}$ are the correct and the incorrect answers, respectively.

\section{Experiments}\label{sec:experiments}

In this section, we describe our experimental setup and then analyze the results of the proposed Procedural Reasoning Networks (PRN) model.

\subsection{Entity Extraction}
Given a recipe, we automatically extract the entities from the initial step of a recipe by using a dictionary of ingredients. While determining the ingredients, we exploit Recipe1M ~\cite{marin2018im2recipe} and Kaggle What’s Cooking Recipes~\cite{yummly2018cooking} datasets, and form our dictionary using the most commonly used ingredients in the training set of RecipeQA. For the cases when no entity can be extracted from the recipe automatically (20 recipes in total), we manually annotate those recipes with the related entities.

\subsection{Training Details}
In our experiments, we separately trained models on each task, as well as we investigated multi-task learning where a single model is trained to solve all these tasks at once. In total, the PRN architecture consists of $\sim$12M trainable parameters. We implemented our models in PyTorch~\cite{paszke2017automatic} using AllenNLP library~\cite{gardner-2018-allennlp}. We used Adam optimizer with a learning rate of 1e-4 with an early stopping criteria with the patience set to 10 indicating that the training procedure ends after 10 iterations if the performance would not improve. We considered a batch size of 32 due to our hardware constraints. In the multi-task setting, batches are sampled round-robin from all tasks, where each batch is solely composed of examples from one task. We performed our experiments on a system containing four NVIDIA GTX-1080Ti GPUs, and training a single model took around 2 hours. We employed the same hyperparameters for all the baseline systems. We plan to share our code and model implementation after the review process.

\subsection{Baselines}\label{sec:baselines}

We compare our model with several baseline models as described below. We note that the results of the first two are previously reported in~\cite{yagcioglu2018recipeqa}.\\ 

\noindent\textbf{Hasty Student}~\cite{yagcioglu2018recipeqa} is a heuristics-based simple model which ignores the recipe and gives an answer by examining only the question and the answer set using distances in the visual feature space.\\
    
\noindent\textbf{Impatient Reader}~\cite{hermann2015teaching} is a simple neural model that takes its name from the fact that it repeatedly computes attention over the recipe after observing each image in the query.\\
    
\noindent\textbf{BiDAF}~\cite{seo2016bidirectional} is a strong reading comprehension model that employs a bi-directional attention flow mechanism to obtain a question-aware representation and bases its predictions on this representation. Originally, it is a span-selection model from the input context. Here, we adapt it to work in a multimodal setting and answer multiple choice questions instead.\\ 
   
\noindent\textbf{BiDAF w/ static memory} is an extended version of the BiDAF model which resembles our proposed PRN model in that it includes a memory unit for the entities. However, it does not make any updates on the memory cells. That is, it uses the static entity embeeddings initialized with GloVe word vectors. We propose this baseline to test the significance of the use of relational memory updates.

\subsection{Results}\label{sec:results}
\begin{figure*}[!t]
\centering
\includegraphics[width=\textwidth]{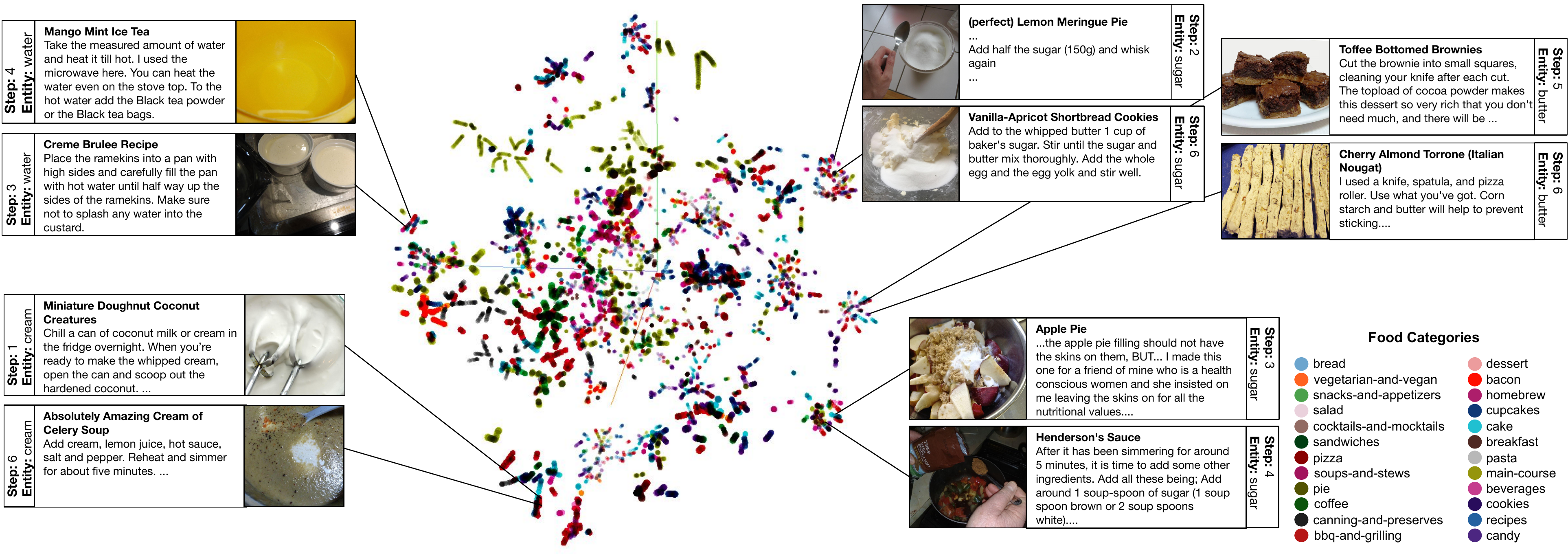}
  \caption{t-SNE visualizations of learned embeddings from each memory snapshot mapping to each entity and their corresponding states from each step for visual cloze task.}
  \label{fig:tsne}
\end{figure*}

\begin{table*}[!t]

\centering
\resizebox{0.99\textwidth}{!}{%
\begin{tabular}{l@{$\quad$}c@{$\;\;$}c@{$\;\;$}c@{$\;\;$}c@{$\quad\quad$}c@{$\;\;$}c@{$\;\;$}c@{$\;\;$}c}
 \toprule
& \multicolumn{4}{c}{Single-task Training} & \multicolumn{4}{c}{Multi-task Training}\\
\cline{2-9}
Model & Cloze & Coherence & Ordering & Average & Cloze & Coherence & Ordering & All\\ 
\hline
Human$^{*}$ & 77.60 & 81.60 & 64.00 & 74.40 & -- & -- & -- & --\\ \hline

Hasty Student & 27.35 & \textbf{65.80} & 40.88 & 44.68 & -- & -- & -- & --\\

Impatient Reader & 27.36 & 28.08 & 26.74 & 27.39 & -- & -- & -- & --\\

BIDAF & 53.95 & 48.82 & 62.42 & 55.06 & 44.62 & 36.00 & \textbf{63.93} & 48.67\\

BIDAF w/ static memory  & 51.82 & 45.88 & 60.90 & 52.87 & \textbf{47.81} & 40.23 & 62.94 & \textbf{50.59}\\

PRN & \textbf{56.31} & 53.64 & \textbf{62.77} & \textbf{57.57} & 46.45 & \textbf{40.58} & 62.67 & 50.17\\
\bottomrule
\multicolumn{8}{l}{\footnotesize{$^{*}$ Taken from the RecipeQA project website, based on 100 questions sampled randomly from the validation set.}} \\

\end{tabular}
}
\caption{Quantitative comparison of the proposed PRN model against the baselines.}
\label{tbl:results}
\end{table*}

Table~\ref{tbl:results} presents the quantitative results for the visual reasoning tasks in RecipeQA. In single-task training setting, PRN gives state-of-the-art results compared to other neural models. Moreover, it achieves the best performance on average. These results demonstrate the importance of having a dynamic memory and keeping track of entities extracted from the recipe. In multi-task training setting where a single model is trained to solve all the tasks at once, PRN and BIDAF w/ static memory perform comparably and give much better results than BIDAF. Note that the model performances in the multi-task training setting are worse than single-task performances. We believe that this is due to the nature of the tasks that some are more difficult than the others. We think that the performance could be improved by employing a carefully selected curriculum strategy~\cite{McCann2018decanlp}.

In Fig.~\ref{fig:tsne}, we illustrate the entity embeddings space by projecting the learned embeddings from the step-by-step memory snapshots through time with t-SNE to 3-d space from 200-d vector space. Color codes denote the categories of the cooking recipes. As can be seen, these step-aware embeddings show clear clustering of these categories. Moreover, within each cluster, the entities are grouped together in terms of their state characteristics. For instance, in the zoomed parts of the figure, chopped and sliced, or stirred and whisked entities are placed close to each other.

\begin{figure*}[!ht]
\centering
\includegraphics[width=0.98\linewidth]{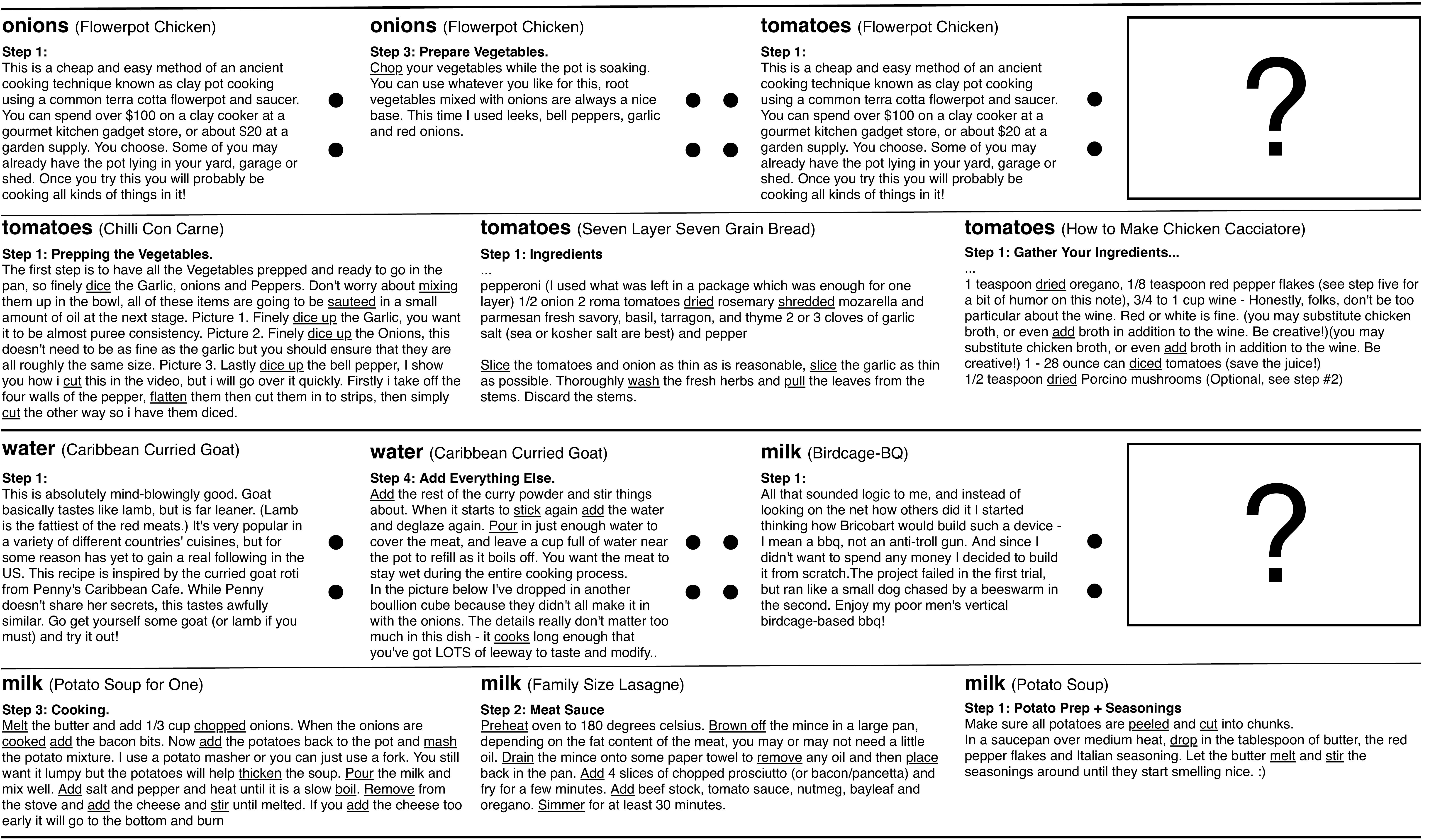}
\caption{Step-aware entity representations can be used to discover the changes occurred in the states of the ingredients between two different recipe steps. The difference vector between two entities can then be added to other entities to find their next states. For instance, in the first example, the difference vector encodes the chopping action done on onions. In the second example, it encodes the pouring action done on the water. When these vectors are added to the representations of raw tomatoes and milk, the three most likely next states capture the semantics of state changes in an accurate manner.} 
\label{fig:nn}
\end{figure*}

Fig.~\ref{fig:nn} demonstrates the entity arithmetics using the learned embeddings from each entity step. Here, we show that the learned embedding from the memory snapshots can effectively capture the contextual information about the entities at each time point in the corresponding step while taking into account of the recipe data. This basic arithmetic operation suggests that the proposed model can successfully capture the semantics of each entity's state in the corresponding step\footnote{We used Gensim for calculating entity arithmetics using cosine distances between entity embeddings.}.

\section{Related Work}\label{sec:related}

In recent years, tracking entities and their state changes have been explored in the literature from a variety of perspectives. In an early work, \citet{henaff2016tracking} proposed a dynamic memory based network which updates entity states using a gating mechanism while reading the text. \citet{bansal2017relnet} presented a more structured memory augmented model which employs memory slots for representing both entities and their relations. \citet{pavez-etal-2018-working} suggested a conceptually similar model in which the pairwise relations between attended memories are utilized to encode the world state. The main difference between our approach and these works is that by utilizing relational memory core units we also allow memories to interact with each other during each update.

\citet{perez-liu-2017-dialog} showed that similar ideas can be used to compile supporting memories in tracking dialogue state. \citet{wang-etal-2017-emergent} has shown the importance of coreference signals for reading comprehension task. More recently, \citet{dhingra2018neural} introduced a specialized recurrent layer which uses coreference annotations for improving reading comprehension tasks. On language modeling task, \citet{ji2017dynamic} proposed a language model which can explicitly incorporate entities while dynamically updating their representations for a variety of tasks such as language modeling, coreference resolution, and entity prediction. 

Our work builds upon and contributes to the growing literature on tracking states changes in procedural text. \citet{bosselut2018simulating} presented a neural model that can learn to explicitly predict state changes of ingredients at different points in a cooking recipe. \citet{mishra2018tracking} proposed another entity-aware model to track entity states in scientific processes. \citet{tandon2018reasoning} demonstrated that the prediction quality can be boosted by including hard and soft constraints to eliminate unlikely or favor probable state changes. In a follow-up work, \citet{du2019consistent} exploited the notion of label consistency in training to enforce similar predictions in similar procedural contexts. \citet{das2018building} proposed a model that dynamically constructs a knowledge graph while reading the procedural text to track the ever-changing entities states. As discussed in the introduction, however, these previous methods use a strong inductive bias and assume that state labels are present during training. In our study, we deliberately focus on unlabeled procedural data and ask the question: Can multimodality help to identify and provide insights to understanding state changes.

\section{Conclusion}\label{sec:conclusion}
We have presented a new neural architecture called Procedural Reasoning Networks (PRN) for multimodal understanding of step-by-step instructions. Our proposed model is based on the successful BiDAF framework but also equipped with an explicit memory unit that provides an implicit mechanism to keep track of the changes in the states of the entities over the course of the procedure. Our experimental analysis on visual reasoning tasks in the RecipeQA dataset shows that the model significantly improves the results of the previous models, indicating that it better understands the procedural text and the accompanying images. Additionally, we carefully analyze our results and find that our approach learns meaningful dynamic representations of entities without any entity-level supervision. Although we achieve state-of-the-art results on RecipeQA, clearly there is still room for improvement compared to human performance. We also believe that the PRN architecture will be of value to other visual and textual sequential reasoning tasks.

\section*{Acknowledgements}
We thank the anonymous reviewers and area chairs for their invaluable feedback. This work was supported by TUBA GEBIP fellowship awarded to E. Erdem; and by the MMVC project via an Institutional Links grant (Project No. 217E054) under the Newton-Katip \c{C}elebi Fund partnership funded by the Scientific and Technological Research Council of Turkey (TUBITAK) and the British Council. We also thank NVIDIA Corporation for the donation of GPUs used in this research. 

\nocite{*}

\bibliography{conll2019}
\bibliographystyle{acl_natbib}

\end{document}